\documentclass[letterpaper, 10 pt, conference]{ieeeconf}  
     
\IEEEoverridecommandlockouts
\usepackage[nolist]{acronym}
\usepackage{xcolor}
\usepackage{amsmath}
\usepackage{amssymb}
\usepackage{mathtools}
\usepackage{mathrsfs}
\usepackage{subcaption}
\usepackage{hyperref}
\usepackage{bm}
\usepackage[noadjust]{cite}
\usepackage{array, makecell} 
\usepackage{tabularx} 
\usepackage{booktabs} 

\usepackage[letterpaper, left=54pt, right=54pt, bottom=54pt, top=54pt]{geometry} 
\usepackage{algpseudocode,algorithm,algorithmicx}
\newcommand*\Let[2]{\State #1 $\gets$ #2}
\algrenewcommand\algorithmicrequire{\textbf{Precondition:}}
\algrenewcommand\algorithmicensure{\textbf{Postcondition:}}

\hypersetup{
    colorlinks,
    citecolor=red,
    filecolor=black,
    linkcolor=blue,
    urlcolor=black
}

\newcommand{\R}{\mathbb{R}}

\DeclarePairedDelimiterX{\inp}[2]{\langle}{\rangle}{#1, #2}

\DeclareMathOperator*{\argmin}{arg\!\,min}

\begin{acronym}[Acronyms]

\acro{DOC}{Direct Optimal Control}
\acro{IOC}{Inverse Optimal Control}
\acro{PIO}{Parametric Inverse Optimization}
\acro{KKT}{Karush-Kuhn-Tucker}
\acro{LMI}{Linear Matrix Inequality}
\acroplural{LMI}[LMIs]{Linear Matrix Inequalities}
\acroindefinite{LMI}{an}{a}

\acro{POP}{Polynomial Optimization Problem}
\acro{SOS}{Sum of Squares}

\acro{PD}{Positive Definite}
\acro{PSD}{Positive Semidefinite}

\acro{LP}{Linear Program}
\acro{QP}{Quadratic Program}
\acro{SDP}{Semidefinite Program}
\acro{QCQP}{Quadratically Constrained Quadratic Programming}

\end{acronym}

\title{\LARGE \bf Reliability of Single-Level Equality-Constrained Inverse Optimal Control}

\author{Filip Be\v{c}anovi\'{c}$^1$, Kosta Jovanovi\'{c}$^1$, Vincent Bonnet$^2$
\thanks{$^1$Filip Be\v{c}anovi\'{c} and Kosta Jovanovi\'{c} are with the School of Electrical Engineering, University of Belgrade, Bul. Kralja Aleksandra 73, 11000 Belgrade, Serbia ({\tt \footnotesize \href{mailto:filip.becanovic@gmail.com}{filip.becanovic@gmail.com}}, {\tt \footnotesize \href{mailto:kostaj@etf.rs}{kostaj@etf.rs}}).}
\thanks{$^2$Vincent Bonnet is with the Laboratory for Analysis and Architecture of Systems, 7 Avenue du Colonel Roche, 31400 Toulouse, France, and with the Image and Pervasive Access Laboratory (IPAL), CNRS-UMI, 2955, Singapore ({\tt \footnotesize \href{mailto:vincent.bonnet@laas.fr}{vincent.bonnet@laas.fr}}).}
\thanks{© 2024 IEEE.  Personal use of this material is permitted.  Permission from IEEE must be obtained for all other uses, in any current or future media, including reprinting/republishing this material for advertising or promotional purposes, creating new collective works, for resale or redistribution to servers or lists, or reuse of any copyrighted component of this work in other works. This is the author's accepted manuscript of an article published in the 2024 IEEE-RAS 23rd International Conference on Humanoid Robots. The final published version is available online at: \href{https://doi.org/10.1109/Humanoids58906.2024.10769923}{https://doi.org/10.1109/Humanoids58906.2024.10769923}.
}
\thanks{Cite as:
F. Bečanović, K. Jovanović and V. Bonnet, "Reliability of Single-Level Equality-Constrained Inverse Optimal Control," 2024 IEEE-RAS 23rd International Conference on Humanoid Robots (Humanoids), Nancy, France, 2024, pp. 623-630.
}
}

\begin{document}

\maketitle
\thispagestyle{empty}
\pagestyle{empty}


\begin{abstract} \label{sec:abstract}
Inverse optimal control (IOC) allows the retrieval of optimal cost function weights, or behavioral parameters, from human motion. The literature on IOC uses methods that are either based on a slow bilevel process or a fast but noise-sensitive minimization of optimality condition violation. Assuming equality-constrained optimal control models of human motion, this article presents a faster but robust approach to solving IOC using a single-level reformulation of the bilevel method and yields equivalent results. Through numerical experiments in simulation, we analyze the robustness to noise of the proposed single-level reformulation to the bilevel IOC formulation with a human-like planar reaching task that is used across recent studies. The approach shows resilience to very large levels of noise and reduces the computation time of the IOC on this task by a factor of 15 when compared to a classical bilevel implementation.
\end{abstract}
\section{Introduction}
\label{sec:introduction}
Human motion generation is a complex task that requires coordination of the human's many degrees of freedom \cite{bernstein1967coordination}. A common hypothesis is that there exists a general principle underlying human motor coordination and that this principle can in essence be represented through an objective function within an Optimal Control Problem (OCP) \cite{sylla2014human}.

Determining the underlying optimal principles was done for specific classes of human motion such as reaching \cite{berret2011evidence,sylla2014human}, grasping \cite{panchea2017inverse}, or gait \cite{clever2016inverse}. For instance, simulation studies have shown that trajectories resulting from minimum-energy optimal control models for human gait coincide well with actual data on human gait \cite{anderson2001dynamic}. For grasping, minimum-angular-jerk or minimum-joint-torque-change optimal control models have been shown to coincide best with data \cite{flash1985coordination, uno1989formation}. 

Numerous, objective functions have been proposed to describe human motion \cite{febrer2023predictive}. However, the idea that a combination of different objectives could be more accurate than any single one in predicting the measured behavior of a system emerged first in the field of reinforcement learning \cite{ng2000algorithms}. It was applied to route planning for outdoor mobile robots, where the desired behavior is often clear, but specifying the reward functions that produce this behavior is much more challenging \cite{ratliff2006maximum}. Within human motion analysis with the model-based paradigm, Mombaur et al. \cite{mombaur2010human} popularized the so-called Inverse Optimal Control (IOC) approach as a way to learn human-like locomotion, but also as a means of transferring said locomotion laws to robots. 

Though many studies focus on determining a stationary \cite{mombaur2010human, berret2011evidence, albrecht2011imitating, albrecht2012bilevel, clever2016inverse, albrecht2017mathematical} and deterministic objective function underlying a particular movement, it is not generally believed that this is how humans function. The objective function may vary through time \cite{lin2016human, jin2019inverse} and may vary across different instances of the motion being performed \cite{westermann2020inverse}. However, studies are focusing on this particular problem due to the extremely time-consuming computational effort associated with identifying even a deterministic and stationary objective that would best fit the measured data \cite{maroger2022inverse}.

The problem of IOC is essentially bilevel: the inner problem represents the optimal-control model of the human motion generation, and the outer problem represents the optimization of the parameters of the motion-generation objective function to minimize differences between the generated and measured motion \cite{mombaur2017inverse, albrecht2017mathematical}.

Bilevel problems are infamously difficult to solve: even linear-linear bilevel problems are NP-hard \cite{dempe2002foundations}. Bilevel problems have existed since the inception of the Stackelberg game in 1936, and can also be expressed as mathematical programs with equilibrium constraints \cite{luo1996mathematical}. Most solution methods focus on more manageable bilevel problems: linear-linear bilevel problems, or bilevel problems where the solution map of the inner problem is a well-defined function, rather than a set map as it is in the general case \cite{bard2013practical}.

The bilevel approach has been criticized in the IOC literature \cite{aghasadeghi2012inverse, johnson2013inverse, panchea2015towards} for taking an extremely long time to compute. An approach, most consistently referred to as Inverse Karush-Kuhn-Tucker (IKKT) \cite{englert2017inverse, englert2018learning}, had become very popular following its publication \cite{keshavarz2011imputing}. IKKT is essentially equivalent to removing the inner loop and putting the norm of the residual of its KKT conditions as the loss function \eqref{eq:ioc-loss}. The IKKT approach provides a polynomial-time algorithm when the data is not noisy and there exists a behavioral parameter, as defined in Section \ref{sec:methods:ioc}, that perfectly describes the system's behavior, \textit{i.e.} when the optimal loss function \eqref{eq:ioc-loss} value is 0. Despite numerous papers using and advocating IKKT \cite{keshavarz2011imputing, aghasadeghi2012inverse, puydupin2012convex, johnson2013inverse, aghasadeghi2014inverse, panchea2015towards, panchea2017inverse, panchea2018human, englert2017inverse, englert2018learning, lin2016human, jin2019inverse, jin2021inverse, westermann2020inverse}, the retrieved optimal weights are rarely used to re-solve the OCP and validate the fitting of human data, with due respect to the exceptions \cite{panchea2018human}.

More recently, studies started showing that IKKT is not robust to noise or modeling error \cite{colombel2022reliability,colombel2023holistic, bevcanovic2022assessing, bevcanovic2022force}. Even a very small amount of noise can lead to a completely different set of optimal recovered weights, as demonstrated practically in the context of IOC with 2 degree-of-freedom planar robot trajectories \cite{colombel2022reliability}. Noise is inherent to human motion and it is now clear in the community that the IKKT approach should not be used as it is formulated currently \cite{colombel2022reliability, colombel2023holistic, bevcanovic2022assessing, bevcanovic2022force}. From a theoretical perspective, the statistical inconsistency of this approach in the presence of noisy data was also proven \cite{aswani2018inverse, aswani2019statistics}.

Research on IKKT drove researchers to examine rank deficiencies in the Jacobian matrix of the KKT conditions with respect to the parameters and dual variables \cite{jin2019inverse, jin2021inverse, colombel2022reliability}, since a rank deficiency would indicate that the IOC problem could be solved. Most recently, a reformulation of the bilevel IOC was proposed using this idea: the inner loop of the bilevel problem is replaced with the condition that the determinant of the Gram matrix of the KKT-Jacobian is equal to zero, which enforces the rank deficiency \cite{colombel2023holistic}. Thus, model weights are removed from the search and the noisy human observations are projected onto the optimal solution set of the OCP. Thereafter, the corresponding OCP solution is input into the IKKT approach to retrieve the set of optimal weights.

Unfortunately, computing the determinant constraint \cite{colombel2023holistic} and its gradient are also time-consuming, forcing them to reduce the size of the inner-loop search space. Doing so requires representing human trajectories as polynomial functions with the smallest degree required to obtain an over-determined system \cite{colombel2023holistic}. So besides computation time, it prevents diverse representations of human trajectories and excludes the use of classical direct-transcriptions of OCPs which lend themselves to efficient solvers \cite{mastalli2020crocoddyl}.

In this context, this paper proposes a single-level reformulation (Section \ref{sec:methods:bilevel}) of the IOC problem that,
while not mathematically equivalent, produces the same solutions as the bilevel approach even in the presence of noisy data
\cite{aswani2018inverse}, but can be solved much more efficiently. The proposed approach was tested in noisy simulations of a 2-degree-of-freedom planar model for a reaching task using human-like cost functions \cite{berret2011evidence} that was also used in recent IOC papers \cite{colombel2022reliability,colombel2023holistic}. 
\section{Methods}
\label{sec:methods}

\subsection{Optimal Control} \label{sec:methods:oc}
A general OCP is given in the Equation \eqref{eq:ocp}. This paper explicitly excludes control limits \eqref{eq:ocp-control-limit} and path constraints \eqref{eq:ocp-path-constraints} as they induce edge cases when computing the sensitivity of the OCP solution with respect to parameters of the OCP, which is a central theme in Section \ref{sec:methods:bilevel}. For conciseness, parameters are not mentioned in Equation \eqref{eq:ocp} but are introduced in Section \ref{sec:methods:ioc}.
\begin{subequations} \label{eq:ocp}
\begin{align} 
    \min_{\bm{x}(.), \bm{u}(.)} \quad & \int_{t_0}^{t_f} \ell(t, \bm{x}(t), \bm{u}(t)) dt && \label{eq:ocp-objective} \\
    \textrm{subject to} \quad & \dot{\bm{x}}(t) = \bm{f}(t, \bm{x}(t), \bm{u}(t)) && \forall t \in [t_0, t_f] \label{eq:ocp-dynamics} \\
    & \bm{b}(\bm{x}(t_0), \bm{x}(t_f)) = 0 && \label{eq:ocp-boundary} \\
    & \bm{u}_{\rm lb} \leq \bm{u}(t) \leq \bm{u}_{\rm ub} && \forall t \in [t_0, t_f] \label{eq:ocp-control-limit} \\
    & \bm{c}(t, \bm{x}(t), \bm{u}(t))\ \leq 0 && \forall t \in [t_0, t_f] \label{eq:ocp-path-constraints}
\end{align}
\end{subequations}

Many methods exist to numerically solve OCPs. Here we opt for the simplest direct transcription using Euler integration, where both the state and control trajectories are considered piece-wise constant,
\begin{subequations}
\begin{align*}
    t_k = t_0 + \frac{k}{N} (t_f - t_0) & \qquad k \in \{ 0 .. N \} \\
    \bm{x}(t) = \bm{x}_k, \quad t \in [t_k, t_{k+1}] & \qquad k \in \{0..N\} \\
    \bm{u}(t) = \bm{u}_k, \quad t \in [t_k, t_{k+1}] & \qquad k \in \{0..N-1\}
\end{align*}
\end{subequations}
and the dynamics equation \eqref{eq:ocp-dynamics} becomes a difference equation
\begin{equation} \label{eq:dynamics-difference}
    \bm{x}_{k+1} = \bm{f}_k(\bm{x}_k, \bm{u}_k), \quad k \in \{0..N-1\}.
\end{equation}

By stacking samples of the states and control in a finite-dimensional vector variable $\bm{z}$, as in \eqref{eq:nlp-variable}, 
\begin{equation} \label{eq:nlp-variable}
    \bm{z} = (\bm{x}_0, \bm{x}_1, \ldots, \bm{x}_N, \bm{u}_0, \bm{u}_1, \ldots \bm{u}_{N-1})^T
\end{equation}
stacking boundary conditions \eqref{eq:ocp-boundary} and the newfound dynamics difference equations \eqref{eq:dynamics-difference} into a vector constraint, as in \eqref{eq:ocp-to-nlp},
\begin{equation} \label{eq:ocp-to-nlp}
    \bm{h}(\bm{z}) = 
    \begin{bmatrix}
        \bm{f}_0(\bm{x}_0, \bm{u}_0) - x_1 \\
        \vdots \\
        \bm{f}_{N-1}(\bm{x}_{N-1}, \bm{u}_{N-1}) - x_N \\
        \bm{b}(\bm{x}_0, \bm{x}_N)
    \end{bmatrix}
\end{equation}
and with a slight abuse of notation reformulating the cost functional \eqref{eq:ocp-objective} into a cost function $f(\bm{z})$, we can reduce the OCP \eqref{eq:ocp} to a Nonlinear Programming (NLP) problem \eqref{eq:nlp} with the finite-dimensional vector variable $\bm{z}$.
\begin{equation} \label{eq:nlp}
\begin{aligned}    
    \min_{\bm{z}} \quad  f(\bm{z}) & \quad    \textrm{subject to} \quad \bm{h}(\bm{z}) = 0
\end{aligned}
\end{equation}
\subsection{Inverse Optimal Control} \label{sec:methods:ioc}
Suppose that there is an NLP \eqref{eq:nlp} that encodes the way a human generates motion, but that $f(\bm{z})$ and possibly $\bm{h}(\bm{z})$ are unknown. As the space of possible functions $f$ and $\bm{h}$ is too large to search through, we must introduce parameters $\bm{\theta}$ that encode this behavior into the functions and then search for the best estimate $\hat{\bm{\theta}}$.

IOC focuses on learning the behavior parameters $\bm{\theta}$ of a system by measuring it. We might have measured the behavior of the system $M$ times, denoting the measured trajectories of this system as $\bm{y}^{(1)}, \bm{y}^{(2)}, \ldots, \bm{y}^{(M)}$, with $\dim{\bm{y}} = \dim{\bm{z}}$. Moreover, our measurements might have different starting and end states encoded in the boundary conditions \eqref{eq:ocp-boundary}, or might in general contain differences in the way the environment is set up. To account for those differences, we introduce a parameter $\bm{x}^{(i)}$, which contains aspects of the environment not already captured in the system's state-control trajectory $\bm{y}^{(i)}$. Be wary of this abuse of notation - $\bm{x}^{(i)}$ is entirely distinct from $\bm{x}(t)$ from Section \ref{sec:methods:oc}.

We adopt this notation so that it coincides with standard notation from the field of Machine Learning (ML), where input-output pairs are often denoted as $(\bm{x}^{(i)}, \bm{y}^{(i)})$, and parameters of the model are denoted with $\bm{\theta}$. The IOC problem is then not distinct from a supervised ML problem. Our NLP system-behavior model \eqref{eq:nlp} now becomes parametrized by behavioral parameters $\bm{\theta}$ and input/environment parameters $\bm{x}$, and our predictions $\hat{\bm{y}}$ become functions of those, as expressed in Equation \eqref{eq:doc}. 
\begin{subequations} \label{eq:doc}
\begin{align}
    \hat{\bm{y}}(\bm{\theta}, \bm{x}) = \argmin_{\bm{z}} \quad & f(\bm{z}; \bm{\theta}, \bm{x}) \label{eq:doc-vars-and-loss} \\
    \textrm{subject to} \quad & \bm{h}(\bm{z}; \bm{\theta}, \bm{x}) = 0 \label{eq:doc-constraints}
\end{align}
\end{subequations}

As is the case in supervised ML, we attempt to learn a mapping between the inputs $\bm{x}$ and outputs $\bm{y}$ from looking at our data points $\left( \bm{x}^{(i)}, \bm{y}^{(i)} \right)_{i=1}^{M}$, by minimizing a cumulative loss function \eqref{eq:ioc-varepsilon} that expresses the discrepancy between the model predictions and the measured data. The sample loss function $\ell$ is often the $L_2$ distance or the RMSE in IOC studies \cite{albrecht2017mathematical, maroger2022inverse}.
\begin{equation} \label{eq:ioc-varepsilon}
    \varepsilon \left( \hat{\bm{y}}^{(1)}, \hat{\bm{y}}^{(2)}, \ldots, \hat{\bm{y}}^{(M)} \right) = \frac{1}{2N} \sum_{i=1}^M \ell(\hat{\bm{y}}^{(i)}, \bm{y}^{(i)})
\end{equation}

The behavioral learning problem that we call IOC is then naturally expressed as the problem in Equation \eqref{eq:ioc}.
\begin{subequations} \label{eq:ioc}
\begin{align}
    \hat{\bm{\theta}} = \argmin_{\bm{\theta}} \quad & \frac{1}{2N} \sum_{i=1}^M \ell(\hat{\bm{y}}^{(i)}, \bm{y}^{(i)}) \label{eq:ioc-loss} \\
    \textrm{subject to} \quad &
    \begin{aligned}[t]
        \hat{\bm{y}}^{(i)} = \argmin_{\bm{z}} \quad & f(\bm{z}; \bm{\theta}, \bm{x}^{(i)}) \\
        \textrm{subject to} \quad & \bm{h}(\bm{z}; \bm{\theta}, \bm{x}^{(i)}) = 0
    \end{aligned} \label{eq:ioc-inner-loop}
\end{align}
\end{subequations}

\subsection{Bilevel Optimization and Single-Level Reformulation} \label{sec:methods:bilevel}
The IOC problem \eqref{eq:ioc} is what we call a bilevel optimization problem. Even the simplest classes of bilevel optimization problems are NP-hard \cite{dempe2002foundations}. Early literature in IOC used this type of formulation and solved the bilevel IOC problem \eqref{eq:ioc} by using a derivative-free solver to minimize the outer-loop cost function \eqref{eq:ioc-loss} while generating model predictions in the inner-loop \eqref{eq:ioc-inner-loop} with a gradient-based solver. This procedure looked somewhat like Algorithm \ref{alg:bilevel-ioc}, where \textsc{solve-ng} is a generic "gradient-less" solver and \textsc{solve} is a generic gradient-based NLP solver.

\begin{algorithm}
    \caption{Bilevel IOC.}
    \label{alg:bilevel-ioc}
    \begin{algorithmic}[1]
        \Require
            \Statex Training set: $\mathcal{D} = ( \bm{x}^{(i)}, \bm{y}^{(i)} )_{i=1}^M$ 
            \Statex Direct model: $f(\bm{z}; \bm{\theta}, \bm{x}), \bm{h}(\bm{z}; \bm{\theta}, \bm{x})$
            \Statex $\bm{z}$-Jacobians: $\partial_{\bm{z}} f(\bm{z}; \bm{\theta}, \bm{x}), \partial_{\bm{z}}  \bm{h}(\bm{z}; \bm{\theta}, \bm{x})$
            \Statex $\bm{z}$-Hessians: $\partial^2_{\bm{z}\bm{z}} f(\bm{z}; \bm{\theta}, \bm{x}), \forall j \; \partial^2_{\bm{z}\bm{z}}  h_j(\bm{z}; \bm{\theta}, \bm{x})$
        \Statex
        \Function{BilevelIO}{$\mathcal{D}$}
            \Let{$\hat{\bm{\theta}}$}{\textsc{solve-ng}(\Call{InnerLoop}{$\bm{\theta}$, $ \mathcal{D}$})}
            \State \Return{$\hat{\bm{\theta}}$}
        \EndFunction
        \Statex
        \Function{InnerLoop}{$\bm{\theta}, \ \mathcal{D}$}
            \Let{$\varepsilon$}{$0$}
            \For{$i \gets 1 \textrm{ to } M$}
                \Let{$\hat{\bm{y}}^{(i)}, \hat{\bm{\nu}}^{(i)}$}{\textsc{solve}($f, \partial_{\bm{z}} f, \partial^2_{\bm{z} \bm{z}} f, \bm{h}, \partial^2_{\bm{z}} \bm{h}, \partial_{\bm{z} \bm{z}} \bm{h}$)}
            \EndFor
            \State \Return{$\varepsilon$}
        \EndFunction
    \end{algorithmic}    
\end{algorithm}

Single-level reformulations of bilevel optimization problems are well-known in the applied mathematics literature \cite{luo1996mathematical, dempe2002foundations, bard2013practical}. The most convenient single-level reformulation for the bilevel IOC problem \eqref{eq:ioc} is the reformulation based on the KKT conditions of the inner-loop \eqref{eq:ioc-inner-loop}, and consists of replacing the inner-loop \eqref{eq:ioc-inner-loop} with its KKT conditions, as it allows the implicit computation of the sensitivity of the OCP solution to the behavioral parameters. If the inner-loop had inequality constraints, \textit{i.e.} contained $\bm{g}(\bm{z}; \bm{\theta}, \bm{x}^{(i)}) \leq 0$, this reformulation would result in a mathematical program with equilibrium constraints \cite{luo1996mathematical, bard2013practical}. Therefore, we conveniently excluded control limits \eqref{eq:ocp-control-limit} and path constraints \eqref{eq:ocp-path-constraints} from consideration in \eqref{eq:ocp}, yielding an NLP \eqref{eq:nlp} and a prediction model \eqref{eq:doc} without inequality constraints.
This is done in almost all the literature on IOC for human motion as it is generally accepted that humans do not operate at their limits during normal motions \cite{panchea2018human}. To establish the KKT conditions of the inner-loop \eqref{eq:ioc-inner-loop}, we go back to the index-free notation of our prediction model \eqref{eq:doc}. The Lagrangian of the prediction model \eqref{eq:doc-lagrangian} requires the introduction of Lagrangian multipliers $\bm{\nu}$ for the equality constraints.
\begin{equation} \label{eq:doc-lagrangian}
    \mathcal{L}(\bm{z}, \bm{\nu}; \bm{\theta}, \bm{x}) = f(\bm{z}; \bm{\theta}, \bm{x}) + \bm{h}(\bm{z}; \bm{\theta}, \bm{x})^T \bm{\nu}
\end{equation}
The first-order necessary conditions of optimality for the prediction model \eqref{eq:doc} with Lagrangian \eqref{eq:doc-lagrangian} are given in equation \eqref{eq:doc-optimality-conditions}.
\begin{subequations} \label{eq:doc-optimality-conditions}
\begin{align}
    \partial_{\bm{z}} \mathcal{L}(\bm{z}, \bm{\nu}; \bm{\theta}, \bm{x})^T &= 0 \\
    \bm{h}(\bm{z}; \bm{\theta}, \bm{x}) &= 0
\end{align}
\end{subequations}
We express this compactly as $\bm{\mathcal{K}}(\bm{z}, \bm{\nu}; \bm{\theta}, \bm{x}) = 0$ where we define the KKT-vector $\bm{\mathcal{K}}$ as in equation \eqref{eq:doc-kkt-vector}.
\begin{equation} \label{eq:doc-kkt-vector}
    \bm{\mathcal{K}}(\bm{z}, \bm{\nu}; \bm{\theta}, \bm{x}) = 
    \begin{bmatrix}
        \partial_{\bm{z}} f(\bm{z}; \bm{\theta}, \bm{x})^T + \partial_{\bm{z}} \bm{h}(\bm{z}; \bm{\theta}, \bm{x})^T \bm{\nu} \\
        \bm{h}(\bm{z}; \bm{\theta}, \bm{x})
    \end{bmatrix}
\end{equation}

By substituting the inner-loop \eqref{eq:ioc-inner-loop} with an analytical condition on its KKT-vector \eqref{eq:doc-kkt-vector} one can obtain a single-level reformulation as in \eqref{eq:single-level-ioc}. Introducing inner-loop Lagrangian multipliers $\bm{\nu}$ and moving them, alongside the inner-loop search variables $\bm{z}$, out to the outer-loop is also required \cite{dempe2002foundations}. The solution to this reformulated IOC \eqref{eq:single-level-ioc} is one where the optimal parameters $\hat{\bm{\theta}}$ produce stationary trajectories $\hat{\bm{y}}^{(i)}$ and multipliers $\hat{\bm{\nu}}^{(i)}$ satisfying KKT conditions \eqref{eq:doc-kkt-vector} of their respective inner loops. This reformulated IOC \eqref{eq:single-level-ioc} is therefore not strictly equivalent to the bilevel IOC \eqref{eq:ioc} \cite{dempe2002foundations}, the justification being related to the fact that the KKT conditions are necessary but not sufficient.

With this in mind, the reformulated single-level IOC variant \eqref{eq:single-level-ioc} searches simultaneously for behavioral parameters $\bm{\theta}$ and corresponding stationary trajectory-multiplier pairs of $\hat{\bm{y}}^{(i)}$ and $\hat{\bm{\nu}}^{(i)}$.

\begin{subequations} \label{eq:single-level-ioc}
\begin{align}
    \left(\hat{\bm{\theta}}, (\hat{\bm{y}}^{(i)}, \hat{\bm{\nu}}^{(i)})_{i=1}^M \right) =& \label{eq:single-level-ioc-solution} \\
    = \argmin_{\bm{\theta}, (\bm{z}^{(i)}, \bm{\nu}^{(i)})_{i=1}^M} \quad & \frac{1}{2N} \sum_{i=1}^M \ell(\bm{z}^{(i)}, \bm{y}^{(i)}) \\
    \textrm{subject to} \quad &
    \bm{\mathcal{K}}(\bm{z}^{(i)}, \bm{\nu}^{(i)}; \bm{\theta}, \bm{x}^{(i)}) = 0 \label{eq:single-level-ioc-KKT}
\end{align}
\end{subequations}
Constraint \eqref{eq:single-level-ioc-KKT} should hold for all examples $i \in \{1..M\}$. 

The KKT-vector can be differentiated and its sensitivity to $\bm{\theta}$, $\bm{z}^{(i)}$, and $\bm{\nu}^{(i)}$  computed and provided to an NLP solver.

The Jacobian $\partial_{\bm{z}, \bm{\nu}} \bm{\mathcal{K}}(\bm{z}, \bm{\nu}; \bm{\theta}, \bm{x})$ of the KKT-vector with respect to primal-dual variables is given in Equation \eqref{eq:doc-kkt-vector-primal-dual-jacobian},
\begin{equation} \label{eq:doc-kkt-vector-primal-dual-jacobian}
    \begin{bmatrix}
        \partial^2_{\bm{z}\bm{z}} f(\bm{z}; \bm{\theta}, \bm{x}) + \sum_j \nu_j \partial^2_{\bm{z}\bm{z}} h_j(\bm{z}; \bm{\theta}, \bm{x}) & \partial_{\bm{z}} \bm{h}(\bm{z}; \bm{\theta}, \bm{x})^T \\
        \partial_{\bm{z}} \bm{h}(\bm{z}; \bm{\theta}, \bm{x}) & 0
    \end{bmatrix}
\end{equation}
while the Jacobian $\partial_{\bm{\theta}} \bm{\mathcal{K}}(\bm{z}, \bm{\nu}; \bm{\theta}, \bm{x})$ of the KKT-vector with respect to behavioral parameters is given in Equation \eqref{eq:doc-kkt-vector-parameter-jacobian}.
\begin{equation} \label{eq:doc-kkt-vector-parameter-jacobian}
    \begin{bmatrix}
        \partial_{\bm{\theta}} (\partial_{\bm{z}} f(\bm{z}; \bm{\theta}, \bm{x}))^T + \sum_j \nu_j \partial_{\bm{\theta}} (\partial_{\bm{z}} h_j(\bm{z}; \bm{\theta}, \bm{x}))^T \\
        \partial_{\bm{\theta}} \bm{h}(\bm{z}; \bm{\theta}, \bm{x})
    \end{bmatrix}
\end{equation}

Using these Jacobians, the single-level IOC procedure reduces to Algorithm \ref{alg:single-level}, where an NLP \eqref{eq:single-level-ioc} is solved with objective $\varepsilon$ and constraints $\bm{\mathcal{K}}$. The algorithm does practically require starting our search from optimal solutions of the inner-loop \eqref{eq:ioc-inner-loop}, so the algorithm contains a call to \textsc{InnerLoop} from Algorithm \ref{alg:bilevel-ioc}.

\begin{algorithm}
    \caption{Single-level IOC with a gradient-based solver.}
    \label{alg:single-level}
    \begin{algorithmic}[1]
        \Require
            \Statex Training set: $\mathcal{D} = ( \bm{x}^{(i)}, \bm{y}^{(i)} )_{i=1}^M$ 
            \Statex Direct model: $f(\bm{z}; \bm{\theta}, \bm{x}), \bm{h}(\bm{z}; \bm{\theta}, \bm{x})$
            \Statex $\bm{z}$-Jacobians: $\partial_{\bm{z}} f(\bm{z}; \bm{\theta}, \bm{x}), \partial_{\bm{z}}  \bm{h}(\bm{z}; \bm{\theta}, \bm{x})$
            \Statex $\bm{z}$-Hessians: $\partial^2_{\bm{z}\bm{z}} f(\bm{z}; \bm{\theta}, \bm{x}), \forall j \; \partial^2_{\bm{z}\bm{z}}  h_j(\bm{z}; \bm{\theta}, \bm{x})$
            \Statex $\bm{\theta}$-Jacobians: $\partial_{\bm{\theta}}  \bm{h}(\bm{z}; \bm{\theta}, \bm{x})$
            \Statex $\bm{\theta}\bm{z}$-Jacobians: $\partial_{\bm{\theta}} (\partial_{\bm{z}} f(\bm{z}; \bm{\theta}, \bm{x}))^T, \forall j \ \partial_{\bm{\theta}} (\partial_{\bm{z}}  h_j(\bm{z}; \bm{\theta}, \bm{x}))^T$
        \Statex
        \Function{SingleLevelIO}{$\bm{\theta_0}$, $\mathcal{D}$}
            \Let{$(\hat{\bm{y}}^{(i)}_0, \hat{\bm{\nu}}^{(i)}_0)_{i=1}^M$}{\Call{InnerLoop}{$\bm{\theta}_0, \ \mathcal{D}$}}
            \Let{$\hat{\bm{\theta}}, \ (\hat{\bm{y}}^{(i)}, \hat{\bm{\nu}}^{(i)})_{i=1}^M$}{\textsc{solve}($\varepsilon, \ \bm{\mathcal{K}}$)}
            \State \Return{$\hat{\bm{\theta}}, \ (\hat{\bm{y}}^{(i)}, \hat{\bm{\nu}}^{(i)})_{i=1}^M$}  
        \EndFunction
    \end{algorithmic}    
\end{algorithm}

One could also use the Jacobians of the KKT-vector, alongside the implicit function theorem, to compute the sensitivity of the OCP's solution $\bm{z}, \bm{\nu}$ with respect to the behavioral parameters $\bm{\theta}$, returning it as a result of the \textsc{InnerLoop} procedure in Algorithm \ref{alg:bilevel-ioc}. Using this result and the chain rule, one could further compute the sensitivity of the bilevel problem's objective function \eqref{eq:ioc-loss} within \textsc{BilevelIO} (Algorithm \ref{alg:bilevel-ioc}), as described in \cite{barratt2018differentiability}. This approach could prove useful in many cases. However, formulating the problem as a simultaneous search over parameters and variables allows the solver to leverage the structure and sparsity of this Jacobian avoiding some potentially costly explicit matrix inversions.
\begin{figure*}[!t] 
    \centering
    \begin{subfigure}[t]{0.32\textwidth}
        \includegraphics[width=\textwidth]{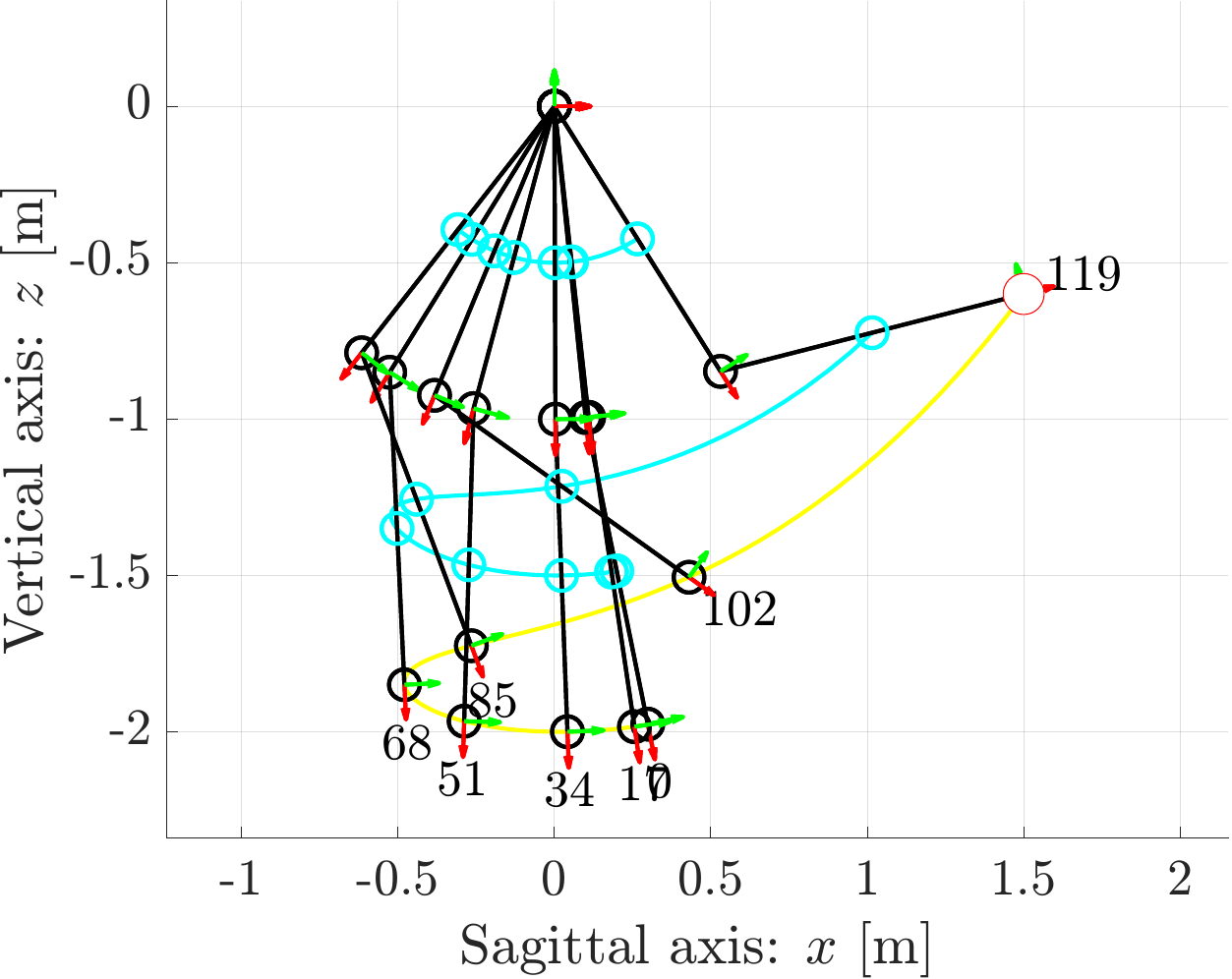}
        \caption{Minimum-torque solution of the robot-arm OCP \textit{i.e.} $\theta_2 = 1, \quad\theta_j = 0, \ j \neq 2$.}
        \label{fig:robot-ocp-results-torque}
    \end{subfigure}
    \hfill
    \begin{subfigure}[t]{0.32\textwidth}
        \includegraphics[width=\textwidth]{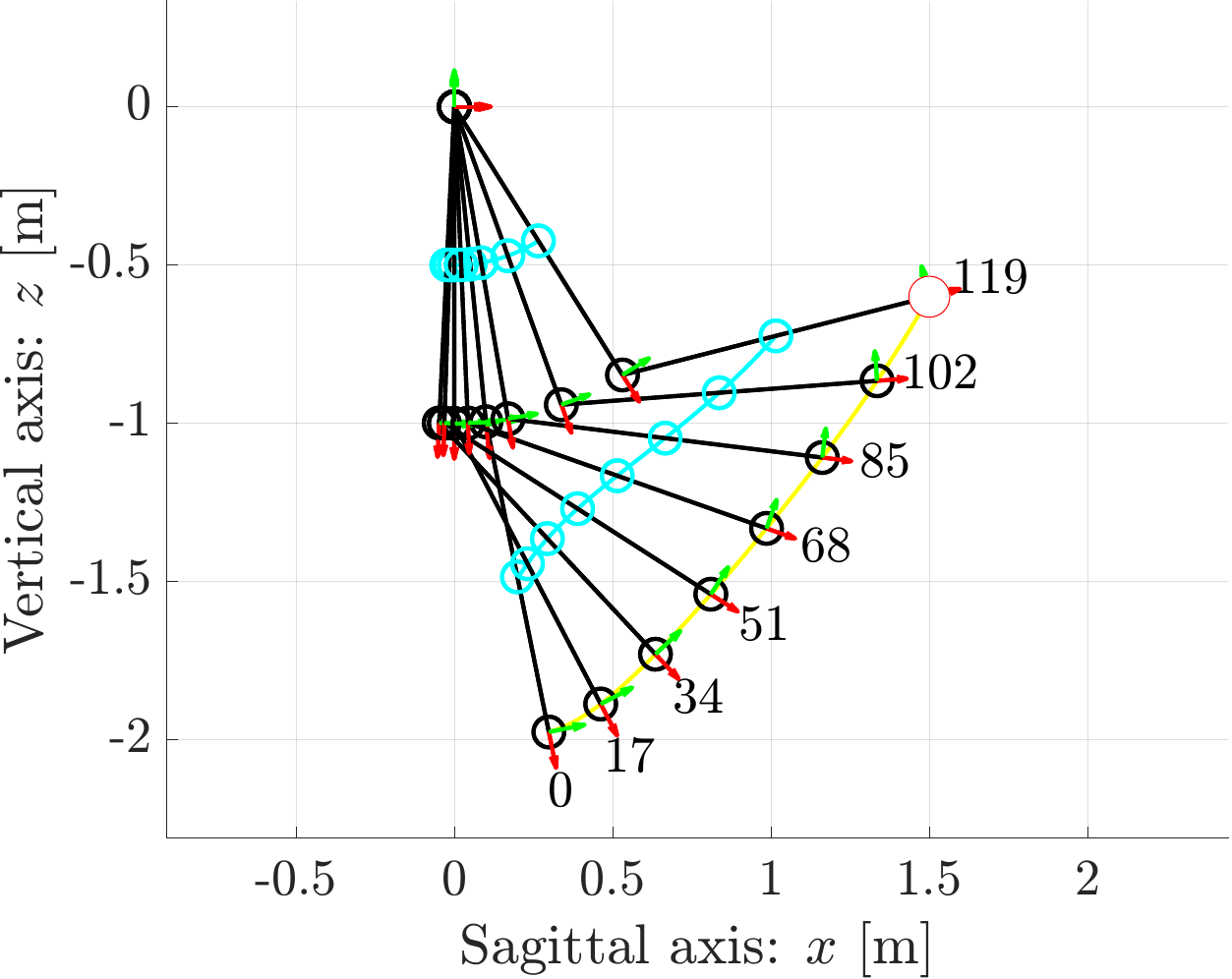}
        \caption{Minimum-end-effector-velocity solution of the robot-arm OCP \textit{i.e.} $\theta_3 = 1, \quad\theta_j = 0, \ j \neq 3$.}
        \label{fig:robot-ocp-results-ee-velocity}
    \end{subfigure}
    \hfill
    \begin{subfigure}[t]{0.32\textwidth}
        \includegraphics[width=\textwidth]{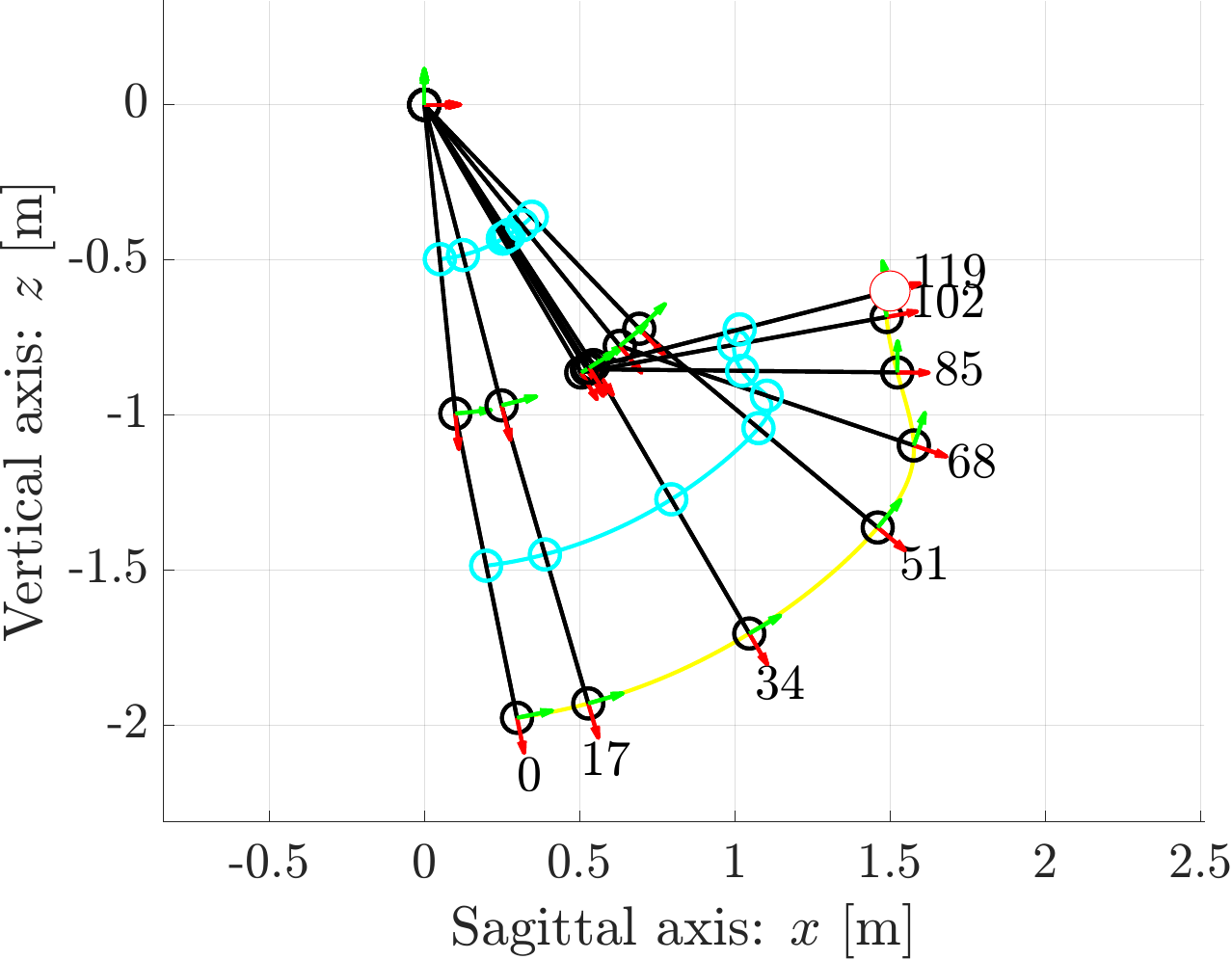}
        \caption{Minimum-torque-change solution of the robot-arm OCP \textit{i.e.} $\theta_5 = 1, \quad\theta_j = 0, \ j \neq 5$.}
        \label{fig:robot-ocp-results-torque-change}
    \end{subfigure}
    \caption{Solutions of the robot-arm OCP \eqref{eq:robot-arm-ocp} with different objective functions. All important constants defined in Section \ref{sec:experiments:robot}.}
    \label{fig:robot-ocp-results}
\end{figure*}
\begin{figure*}[!t]
    \centering
    \begin{subfigure}[t]{0.47\textwidth}
        \centering
        \includegraphics[width=\textwidth]{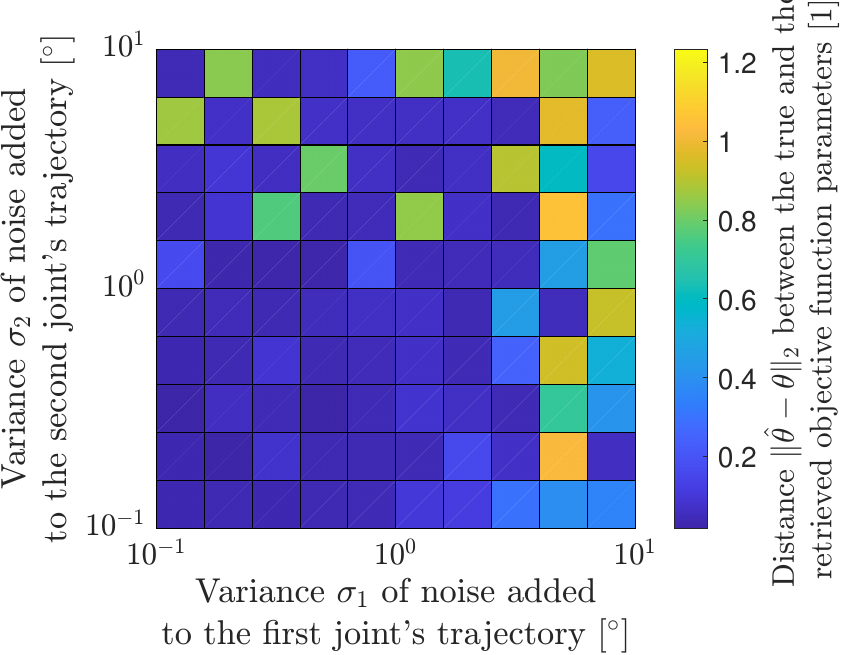}
        \caption{Heatmap of the error in behavioral parameter retrieval.}
        \label{fig:noise-heatmap-parameters}
    \end{subfigure}
    \hfill
    \begin{subfigure}[t]{0.47\textwidth}
        \centering
        \includegraphics[width=\textwidth]{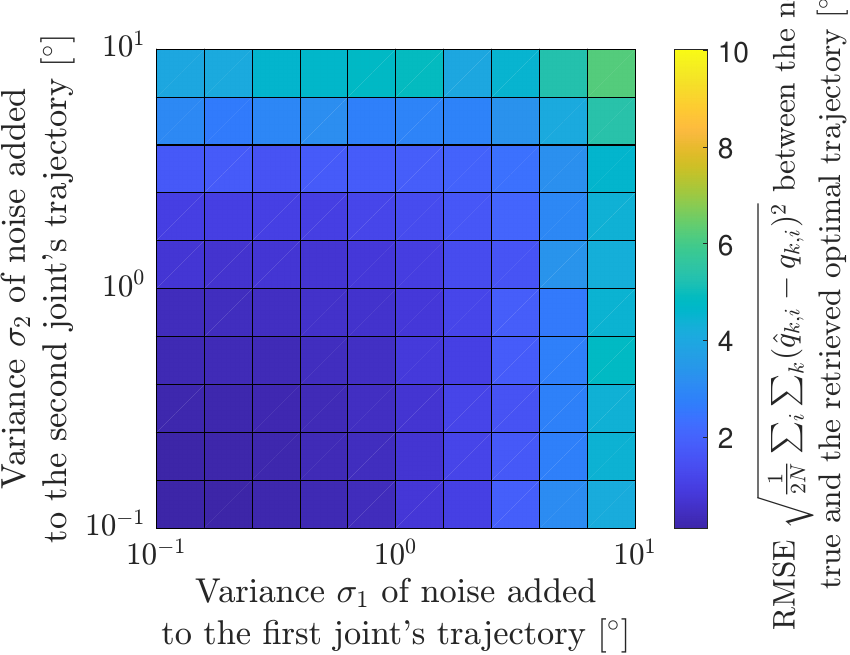}
        \caption{Heatmap of the error in joint trajectory retrieval.}
        \label{fig:noise-heatmap-trajectory}
    \end{subfigure}
    \caption{Heatmaps of the errors in parameter and trajectory retrieval when solving the single-level IOC problem \eqref{eq:single-level-ioc} with noisy data using Algorithm \ref{alg:single-level}.}
    \label{fig:noise-heatmap}
\end{figure*}
\subsection{Optimal Control of Robot Arm Trajectory} \label{sec:methods:robot}
For a robotic or human arm, the dynamics are expressed in the space of the generalized coordinates, \textit{i.e.}, joint angles $\bm{q}$, and relate the generalized forces, \textit{i.e.}, joint torques $\bm{\tau}$, with the generalized coordinates $\bm{q}$, velocities $\dot{\bm{q}}$, and accelerations $\ddot{\bm{q}}$. These dynamics are usually described by the equation given in \eqref{eq:robot-dynamics},
\begin{equation} \label{eq:robot-dynamics} 
    \bm{\tau}_k = \bm{M}(\bm{q}_k) \ddot{\bm{q}}_k + \bm{C}(\bm{q}_k, \dot{\bm{q}}_k) + \bm{G}(\bm{q}_k)
\end{equation}
where $\bm{M}(\bm{q}_k)$ is the configuration-dependent mass matrix, $\bm{C}(\bm{q}_k, \dot{\bm{q}}_k)$ is the vector of centrifugal and Coriolis forces, and $\bm{G}(\bm{q}_k)$ is the vector of gravitational forces.

For simplicity's sake, we will equivalently treat the arm as a double integrator like in Equation \eqref{eq:double-integrator}, with $\bm{\tau}_k$ treated as a non-linear function of the state rather than the control signal.
\begin{subequations} \label{eq:double-integrator}
\begin{align}
    \bm{q}_{k+1} &= \bm{q}_k + \Delta t \, \dot{\bm{q}}_k \\
    \dot{\bm{q}}_{k+1} &= \dot{\bm{q}}_k + \Delta t \, \ddot{\bm{q}}_k
\end{align}
\end{subequations}

Throughout the paper, we focus on an OCP whose goal is to move the arm from an initial joint configuration $ \bm{q}_i \in \R^n $, towards an end-effector goal location $ \bm{p}_g \in \R^2 $.

\begin{table}[t]
    \centering
    \renewcommand{\arraystretch}{1.5}
    \caption{Objective functions considered in this study \cite{berret2011evidence}.}
    \begin{tabular}{|ll|}
        \hline
        Objective Function & Mathematical Expression \\
        \hline
        Min. joint velocities & $\phi_1 = \sum_{k=0}^{N-2} \dot{\bm{q}}_k^T \dot{\bm{q}}_k$ \\
        Min. joint torques & $\phi_2 = \sum_{k=0}^{N-3} \bm{\tau}_k^T \bm{\tau}_k$ \\
        Min. end-effector velocity & $\phi_3 = \sum_{k=0}^{N-2} \bm{v}_t^T \bm{v}_t$ \\
        Min. joint jerk & $\phi_4 = \sum_{k=0}^{N-2} \dddot{\bm{q}}_k^T \dddot{\bm{q}}_k$ \\
        Min. joint torque changes & $\phi_5 = \sum_{k=0}^{N-3} \dot{\bm{\tau}}_k^T \dot{\bm{\tau}}_k$ \\
        \hline
    \end{tabular}
    \label{tab:robot-arm-ocp-basis}
\end{table}

The OCP searches for discretized joint positions $\mathbf{q}$, velocities  $\mathbf{\dot{q}}$, and accelerations $\mathbf{ \ddot{q}}$, that satisfy Euler integration constraints \eqref{eq:double-integrator}, \textit{i.e.} \eqref{eq:robot-arm-ocp-euler-q}-\eqref{eq:robot-arm-ocp-euler-dq}, initial and terminal conditions \eqref{eq:robot-arm-ocp-initial}-\eqref{eq:robot-arm-ocp-goal}, and minimize some compound objective function \eqref{eq:robot-arm-ocp-objective} containing $\bm{q}$, $\dot{\bm{q}}$, $\ddot{\bm{q}}$, $\bm{\tau}$ and possibly other derived quantities.
\begin{subequations} \label{eq:robot-arm-ocp}
\begin{align}
    \min_{\bm{q}, \dot{\bm{q}}, \ddot{\bm{q}}} \quad & \sum_{j=1}^r  \theta_j \phi_{j}(\bm{q}, \dot{\bm{q}}, \ddot{\bm{q}}) \label{eq:robot-arm-ocp-objective} \\
    \textrm{subject to} \quad & 
    \bm{q}_{k+1} = \bm{q}_k + \Delta t \, \dot{\bm{q}}_k \quad k \in \{ 0..N-1 \} \label{eq:robot-arm-ocp-euler-q} \\
    & \dot{\bm{q}}_{k+1} = \dot{\bm{q}}_k + \Delta t \, \ddot{\bm{q}}_k \quad k \in \{ 0..N-2 \} \label{eq:robot-arm-ocp-euler-dq} \\
    & \bm{q}_0 = \bm{q}_i \label{eq:robot-arm-ocp-initial} \\
    & \textrm{FK}(\bm{q}_{N}) = \bm{p}_g \label{eq:robot-arm-ocp-goal} 
\end{align}
\end{subequations}

This setup maps over naturally to our general description of the prediction model \eqref{eq:doc} in Section \ref{sec:methods:ioc} and the KKT-vector \eqref{eq:doc-kkt-vector}
in Section \ref{sec:methods:bilevel} by observing the transformations in Equation \eqref{eq:robot-to-ioc-notation},
\begin{subequations} \label{eq:robot-to-ioc-notation}
\begin{gather}
    \bm{z} = (\bm{q}, \dot{\bm{q}}, \ddot{\bm{q}}) \label{eq:robot-to-ioc-notation-z} \\
    \bm{x} = (\bm{q}_i, \bm{p}_e) \label{eq:robot-to-ioc-notation-x} \\
    \bm{\theta} = (\theta_1, \ldots, \theta_r) \label{eq:robot-to-ioc-notation-theta}\\
    f(\bm{z}; \bm{\theta}) = \sum_{j=1}^r  \theta_j \phi_{j}(\bm{q}, \dot{\bm{q}}, \ddot{\bm{q}}) \label{eq:robot-to-ioc-notation-f}\\
    \bm{h}(\bm{z}; \bm{x}) = (\ldots, \bm{q}_0 - \bm{q}_i, \textrm{FK}(\bm{q}_{N}) - \bm{p}_g) \label{eq:robot-to-ioc-notation-h}
\end{gather}
\end{subequations}
where the dots in \eqref{eq:robot-to-ioc-notation-h} stand for the stacking of $\bm{q}_{k+1} - \bm{q}_k + \Delta t \, \dot{\bm{q}}_k$ for \eqref{eq:robot-arm-ocp-euler-q} and $\dot{\bm{q}}_{k+1} - \dot{\bm{q}}_k + \Delta t \, \ddot{\bm{q}}_k$ for \eqref{eq:robot-arm-ocp-euler-dq}.

Note that the predictive model objective function in \eqref{eq:robot-to-ioc-notation-f} is not a function of the environment parameters $\bm{x}$, and that the equality constraints of the predictive model in \eqref{eq:robot-to-ioc-notation-h} are not functions of the behavioral parameters $\bm{\theta}$. This setup better reflects the majority of studies in IOC that consider the behavioral parameters $\theta_j$ only in the objective function, and in particular as weights of basis objective functions $\bm{\phi}_j$ \cite{bevcanovic2022force, bevcanovic2022assessing}.

Many basis objective functions can be found throughout the robotics and biomechanics literature on IOC. Table \ref{tab:robot-arm-ocp-basis} uses five of the cost functions proposed by Berret et al. \cite{berret2011evidence} for a similar reaching task.
\section{Numerical Experiments} \label{sec:experiments}

\begin{figure*}[h]
    \centering
    \begin{subfigure}[t]{0.46\textwidth}
        \centering
        \includegraphics[width=\textwidth]{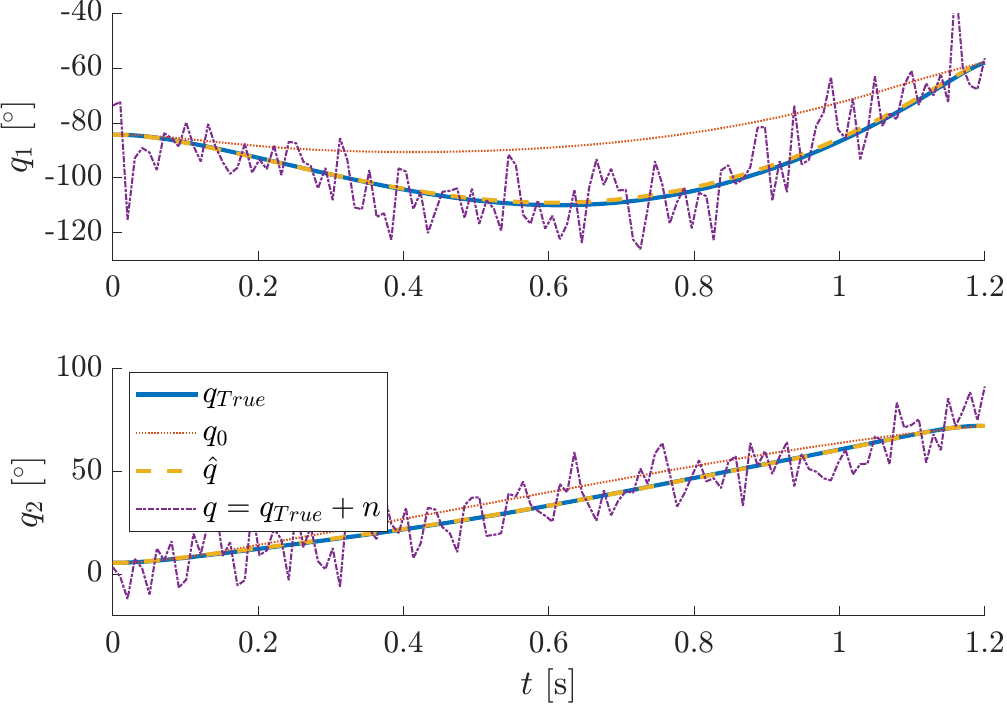}
        \caption{Joint trajectories with the noise level that yielded the maximum trajectory RMSE in the heatmap of Figure \ref{fig:noise-heatmap-trajectory}, \textit{i.e.} $\sigma_1 = 10^1$, $\sigma_2 = 10^1$, $\textrm{RMSE}(\bm{q}, \hat{\bm{q}}) = 10.03^\circ$.}
        \label{fig:joint-trajectories-max-noise}
    \end{subfigure}
    \hfill
    \begin{subfigure}[t]{0.46\textwidth}
        \centering
        \includegraphics[width=\textwidth]{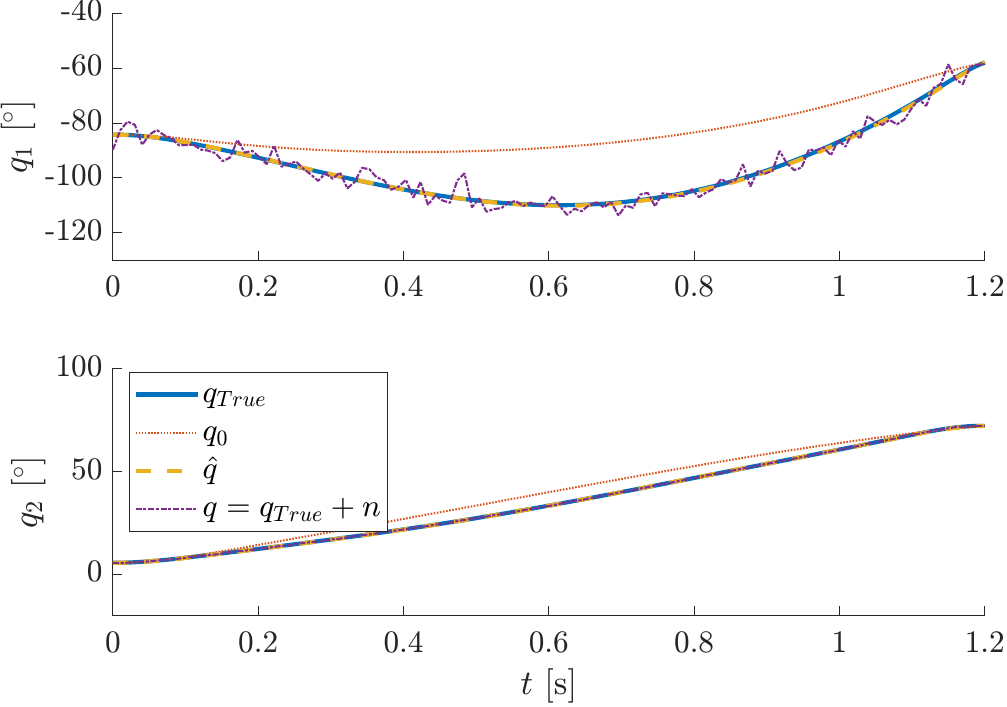}
        \caption{Joint trajectories with the noise level that yielded the median trajectory RMSE in the heatmap of Figure \ref{fig:noise-heatmap-trajectory}, \textit{i.e.} $\sigma_1 = 10^{0.4}$, $\sigma_2 = 10^{-1}$, $\textrm{RMSE}(\bm{q}, \hat{\bm{q}}) = 2.37^\circ$.}
        \label{fig:joint-trajectories-med-noise}
    \end{subfigure}
    \caption{Joint trajectories involved in the noisy IOC from Section \ref{sec:experiments:ioc} and the heatmaps from Figure \ref{fig:noise-heatmap} for two different levels of noise. }
    \label{fig:joint-trajectories}
\end{figure*}
\subsection{Optimal Control of Robot Arm Trajectory} \label{sec:experiments:robot}
The 2 degrees-of-freedom arm should move from its given initial configuration $\bm{q}_i = \left(-\frac{\pi}{2} + 0.1, -0.1\right) \, \text{rad}$, analogous to slightly flexed arm hanging from the shoulder, towards a given final configuration $\bm{p}_e = (1.5, 0.6) \, \text{m}$, analogous to a position in front of the body where a grasping target may be. The trajectory starts at $t_0 = 0 \, \text{s}$ and lasts a fixed amount until $t_f = 1.2 \, \text{s}$. The trajectory is discretized with $N = 120$ samples, so the time-step is equal to $\Delta t = 0.01 \, \text{s}$.

For simplicity, the segments are chosen to be unit-length, unit-mass, uniform-density sticks, so that the geometric and inertial parameters are as simple as possible: the segment lengths $L = (1, 1) \, \text{m}$, center-of-mass local-frame positions $C = (0.5, 0.5; 0, 0) \, \text{m}$, segment masses $M = (1, 1) \, \text{kg}$, and segment moments of inertia computed at the axis perpendicular to the plane of motion and passing through the center of mass $I = \left( \frac{1}{12}, \frac{1}{12} \right) \, \text{kg} \cdot \text{m}^2$. 
The gravitational acceleration is fixed to $9.81 \, \text{m} \cdot \text{s}^{-2}$.

In Figure \ref{fig:robot-ocp-results}, the results of solving the OCP given in problem \eqref{eq:robot-arm-ocp} for three different values of $\theta_j$ are displayed. Multiple snapshots of the robot arm are shown in various configurations along the trajectory, with sample numbers ranging from 0 to 119. The robot body is drawn in black, while the $XY$-axes of the reference frames of the robot's joints are represented as red-green arrows. The centers of mass of the segments are depicted as cyan circles, with their trajectory indicated by a cyan line. Finally, the trajectory of the end-effector is shown as a yellow line.

Subfigure \ref{fig:robot-ocp-results-torque} represents the minimum-torque trajectory. One can see that the trajectory swings back using gravity to build up acceleration, saving up joint torques. Moreover, the centers of mass of the links are brought closer to the vertical line passing through the shoulder joint, aligning the shoulder-to-center of mass line with the gravitational force. This alignment diminishes the torque required to compensate for gravity. 
Subfigure \ref{fig:robot-ocp-results-ee-velocity} represents the minimum sum of end-effector velocity trajectory. One can see that the end-effector travels from its initial position to its goal position in a straight line.
Subfigure \ref{fig:robot-ocp-results-torque-change} represents the minimum-torque-change trajectory. This trajectory does not produce torque oscillations like the minimum-torque trajectory, while still preventing significant increments in torque usage.
\subsection{Noisy Inverse Optimal Control} \label{sec:experiments:ioc}
With a given behavioral parameter, and other constants fixed as in Section \ref{sec:experiments:robot}, we generate an optimal solution $\bm{y}^{(1)}$ for the corresponding $\bm{x}^{(1)}$ (forming $\bm{x}^{(1)}$ described in \eqref{eq:robot-to-ioc-notation-x}, numerical values in Section \ref{sec:experiments:robot}). Select $N_{noise} = 11$ logarithmically-spaced joint-dependent Gaussian-noise levels and add it to each joint's trajectory. For each noise level, process a single-example data set $(\bm{x}^{(1)}, \bm{y}^{(1)})$ with Algorithm \ref{alg:single-level}, using a random initial behavioral parameter $\bm{\theta}_0$. Obtain a solution to the IOC problem \eqref{eq:single-level-ioc} as an estimate $\hat{\bm{\theta}}$ of the behavioral parameters and $\hat{\bm{y}}^{(1)}$ of the measured trajectory. 

The IOC loss function \eqref{eq:ioc-loss} used in this numerical experiment is the quadratic loss in the joint angle space, \textit{i.e.}
\begin{equation}
    \ell(\hat{\bm{y}}^{(i)}, \bm{y}^{(i)}) = \| \hat{\bm{q}}^{(i)} - \bm{q}^{(i)} \|_2^2.
\end{equation}
One can extract $\bm{q}$ from $\bm{y}$ as described in \eqref{eq:robot-to-ioc-notation-z}.

Figure \ref{fig:noise-heatmap} displays the results of this analysis. The $X$-axes of the figures represent the variance of the Gaussian noise added to the trajectory of joint 1, ranging logarithmically from $0.1^\circ$ to $10^\circ$. The $Y$-axes have the same range but represent the variance of Gaussian noise added to the trajectory of joint 2. Both heatmaps represent measures of the intensity of errors.

In Subfigure \ref{fig:noise-heatmap-parameters}, the heatmap represents $L_2$ distance $\|\hat{\bm{\theta}} - \bm{\theta} \|$ between the true behavioral parameter $\bm{\theta}$ and the estimated one $\hat{\bm{\theta}}$ given the different noise levels, and it is unitless.
In Subfigure \ref{fig:noise-heatmap-trajectory}, the heatmap represents the RMSE, in degrees, between the true trajectory $\bm{q}^{(1)}$ and the retrieved trajectory $\hat{\bm{q}}$ given the different noise levels. 
From Subfigure \ref{fig:noise-heatmap-parameters}, one can see that few noise levels produce a significant error in behavioral parameter retrieval.
From Subfigure \ref{fig:noise-heatmap-trajectory}, one can see that as the noise levels increase, the error in trajectory retrieval increases linearly.

It is possible to see from comparing Subfigure \ref{fig:noise-heatmap-parameters} to Subfigure \ref{fig:noise-heatmap-trajectory} that for some noise levels the behavioral parameter solving the problem is not unique: the behavioral parameters do not match well but the trajectories do. We attribute this to the poor sensitivity of the optimal solution with respect to parameters when correlated functions are present, \textit{e.g.} joint velocity and joint jerk. This means that in this region of the behavioral parameter space, varying the parameters does not produce much of a change in the optimal solution, but this small change still fits the noisy data slightly better. This can be considered a case of what is called overfitting in ML. To prevent this, we should train with more than one sample, which we plan to do in future works.

Figure \ref{fig:joint-trajectories} illustrates the joint trajectories of the noisy IOC for two different levels of noise. The full line $\bm{q}_{True}$ represents the true target trajectory generated with a given behavioral parameter $\bm{\theta}_{True}$. The dotted line $\bm{q}_0$ represents the trajectory corresponding to behavioral parameter $\bm{\theta}_0$ with which we initialize the single-level IOC search, like in Algorithm \ref{alg:single-level}. The dashed-dotted line $\bm{q}$ is the noisy target trajectory which we feed into the loss function \eqref{eq:ioc-loss}, obtained from the true target trajectory by adding noise. The dashed line $\hat{\bm{q}}$ represents the recovered trajectory from the single-level IOC procedure with Algorithm \ref{alg:single-level}. Figure \ref{fig:joint-trajectories} is meant to complement Figure \ref{fig:noise-heatmap} and strenghten the point that even in the presence of heavy noise, we are able to retrieve a good fit to the trajectory.

As far as computation time goes, across all noise levels, Algorithm \ref{alg:bilevel-ioc} averages $8.240 \ \text{s}$ to compute the solution while Algorithm \ref{alg:single-level} averages $0.549 \ \text{s}$ to compute the solution. The computation time is improved by a factor of 15 with respect to a standard bilevel formulation.
In this context, we can conclude that the single-level IOC is fairly robust to noise, contrary to what we referred to earlier as the IKKT procedure \cite{keshavarz2011imputing} which has been shown to have large behavioral parameter and trajectory errors for noise levels equal to $0.1^\circ$ \cite{colombel2022reliability}.
\section{Conclusion}
\label{sec:conclusion}
The objective of this paper was to demonstrate that single-level reformulations of the bi-level IOC can work in the context of noisy observations while being considerably faster. Moreover, it allows for more general formulations than previous work  \cite{colombel2023holistic} and is robust to noise contrary to the IKKT method \cite{jin2019inverse, westermann2020inverse}. Though validated in a simulation, the proposed approach is encouraging, demonstrating robustness to substantial amounts of noise. The requirement for using this approach, provided the motion-generation OCP model is only equality-constrained, is the $C^1$ continuity of the motion-generation objective and constraint functions with respect to both the trajectory variables and the behavioral parameters. The huge computational speed-up comes from the fact that the NLP solver in Algorithm \ref{alg:single-level} can implicitly compute the sensitivity of local solutions of the motion-generation problem with respect to the behavioral parameters and thus reduce the bilevel to a single-level problem \cite{barratt2018differentiability}. 

This study doesn't consider control problems where the OCP contains inequality constraints, because they are more difficult to handle mathematically, but they represent one of the main advantages of OCP-based planning. This is a legitimate concern and efforts should be made to use single-level reformulations and algorithms for IOC with inequality-constrained OCPs \cite{albrecht2017mathematical}. Nevertheless, when comparing experimental human motion data to models of the corresponding motion planning problem, the data is often far from activating the inequality constraints that are reasonably applicable in the model \cite{panchea2018human}. Therefore, we believe this type of single-level IOC could be used as-is in simple human motion analysis applications.

Single-level reformulations of bilevel problems, that are based on KKT conditions, always suffer the same setback when treating non-convex inner problems, \textit{i.e.}, the KKT conditions are only necessary and not sufficient. The method in this study is no exception, and the stationary trajectories it finds \eqref{eq:single-level-ioc-solution} may be saddle-points or maximizers of the inner problem, rather than minimizers \cite{dempe2002foundations}. Such effects did not arise in numerical simulations performed in this study, but this does not prohibit them from emerging as more complex OCP problems are considered. Nevertheless, when comparing experimental human motion data to models of the corresponding motion planning problem, the data tends to be closer to minimizers, which may help in guiding the single-level IOC to correct solutions \cite{berret2011evidence, sylla2014human, mombaur2010human, clever2016inverse, albrecht2011imitating, albrecht2012bilevel, albrecht2017mathematical, lin2016human, jin2019inverse, maroger2022inverse, mombaur2017inverse, aghasadeghi2012inverse, puydupin2012convex, panchea2018human, westermann2020inverse}.

The numerical simulations in this paper are simple. The plant dynamics \eqref{eq:robot-arm-ocp-euler-q}-\eqref{eq:robot-arm-ocp-initial} are linear, a single constraint \eqref{eq:robot-arm-ocp-goal} is nonlinear, most (3 out of 5) objective functions \eqref{eq:robot-arm-ocp-objective} considered in Table \ref{tab:robot-arm-ocp-basis} are convex. The results should be considered preliminary, yet they show promise. Of interest remains the question of whether this IOC approach will be viable or will its shortcomings emerge with larger and more complex systems being considered, \textit{e.g.}, underactuated systems with many degrees of freedom and nonlinear constraints.

Finally, studying the performance of single-level IOC can be extended along multiple axes, most notably for 1) other transcriptions of the OCP than the direct-Euler transcription, 2) other task formulations than the one where the final position of the end-effector is constrained, 3) more extensive sets of basis objective functions than given in \ref{tab:robot-arm-ocp-basis}, 4) the use of variable cost function weights as the fast computation time will allow it, and of course 5) the use of real human data.
\section*{Acknowledgement}

The Ministry of Science, Technological Development and Innovation of the Republic of Serbia financially supported this work under contract number 451-03-66/2024-03/200103.
This work was financially supported by the European Union’s Horizon Europe Programme for Research and Innovation through the MUSAE project with grant agreement N\textsuperscript{o} 101070421.
This work was supported by the Partenariats Hubert Curien (PHC) Pavle Savi\'{c}, funded by the French Ministry of Europe and Foreign Affairs and the French Ministry of Higher Education, Research, and Innovation under grant agreement N\textsuperscript{o} 49431WM.


\bibliographystyle{IEEEtran}
\bibliography{references.bib}

\end{document}